\documentclass[10pt,twocolumn]{article}

\usepackage{authblk}

\makeatletter
\renewcommand{\maketitle}{\bgroup\setlength{\parindent}{0pt}
\begin{flushleft}
  \textbf{\@title}

  \@author
\end{flushleft}\egroup
}
\makeatother

\title{\huge Ground-Truthing AI Energy Consumption: Validating CodeCarbon Against External Measurements  \vspace{0.5cm}}
\author{\Large Raphael Fischer \normalsize \hspace{0.3cm} Lamarr Institute, TU Dortmund University \hspace{0.3cm} \emph{raphael.fischer@tu-dortmund.de}}

\date{\today}

\usepackage[showframe=false, margin=.75in]{geometry}

\usepackage{abstract}
\renewcommand{\abstractname}{}    

\usepackage{afterpage}

\renewenvironment{abstract}
 {\small
  \begin{center}
  \bfseries \abstractname\vspace{-.5em}\vspace{0pt}
  \end{center}
  \list{}{%
    \setlength{\leftmargin}{10mm}
    \setlength{\rightmargin}{\leftmargin}%
  }%
  \item\relax}
 {\endlist}

\usepackage[utf8]{inputenc}

\usepackage{wrapfig}
\usepackage[lofdepth,lotdepth]{subfig}
\usepackage{booktabs}
\usepackage{rotating}
\usepackage{caption}
\usepackage{csquotes}
\usepackage{todonotes}
\usepackage{stfloats}
\usepackage[T1]{fontenc}
\usepackage{textcomp}
\usepackage{times}

\usepackage{framed} 

\usepackage[]{biblatex}
	\renewbibmacro*{volume+number+eid}{%
  \printfield{volume}%
  \setunit*{\addnbspace}
  \printfield{number}%
  \setunit{\addcomma\space}%
  \printfield{eid}}
\DeclareFieldFormat[article]{number}{\mkbibparens{#1}}
\usepackage{setspace}
\usepackage{color}
\definecolor{black}{gray}{0} 
\usepackage[colorlinks=true,linkcolor=black,citecolor=black]{hyperref}

\usepackage{tabularx}
\newcolumntype{b}{X}
\newcolumntype{s}{>{\hsize=.5\hsize}X}

\usepackage{ntheorem}

\begin{document}

\twocolumn[
\begin{@twocolumnfalse}
\maketitle
\begin{abstract}
\textit{Although machine learning (ML) and artificial intelligence (AI) present fascinating opportunities for innovation, their rapid development is also significantly impacting our environment.
In response to growing resource-awareness in the field, quantification tools such as the \emph{ML Emissions Calculator} and \emph{CodeCarbon} were developed to estimate the energy consumption and carbon emissions of running AI models.
They are easy to incorporate into AI projects, however also make pragmatic assumptions and neglect important factors, raising the question of estimation accuracy.
This study systematically evaluates the reliability of static and dynamic energy estimation approaches through comparisons with ground-truth measurements across hundreds of AI experiments.
Based on the proposed validation framework, investigative insights into AI energy demand and estimation inaccuracies are provided.
While generally following the patterns of AI energy consumption, the established estimation approaches are shown to consistently make errors of up to 40\%.
By providing empirical evidence on energy estimation quality and errors, this study establishes transparency and validates widely used tools for sustainable AI development.
It moreover formulates guidelines for improving the state-of-the-art and offers code for extending the validation to other domains and tools, thus making important contributions to resource-aware ML and AI sustainability research.
\bigskip}
\end{abstract}
\end{@twocolumnfalse}
]

\section{Introduction}

Artificial intelligence (AI) holds many promises for our world, however does not come without costs.
While modern AI enables sustainable innovation in domains such as construction, transportation, healthcare, and manufacturing~\cite{kar_how_2022}, it also poses critical risks to our society, economy, and environment---or, in other words, across all sustainability dimensions~\cite{van_wynsberghe_sustainable_2021,fischer_diss}.
The particular danger of ``systemic risks'' and resulting need for ``environmental sustainability'' was manifested by experts worldwide, for example in the \emph{International AI Safety Report}~\cite{bengio_international_2025}.
Unfortunately, the field of machine learning (ML) lacks resource-awareness and is mostly focused on boosting predictive performance~\cite{fischer_towards_2024}.
Given that publications often fail to justify their compute expenses or discuss potential negative consequences~\cite{birhane_values_2022}, the observable ``compute trends''~\cite{9891914_compute_trends} and ``bigger-is-better paradigm''~\cite{varoquaux_hype_2024} are concerning but not surprising.

\begin{figure}
    \centering
    \includegraphics[width=.84\linewidth]{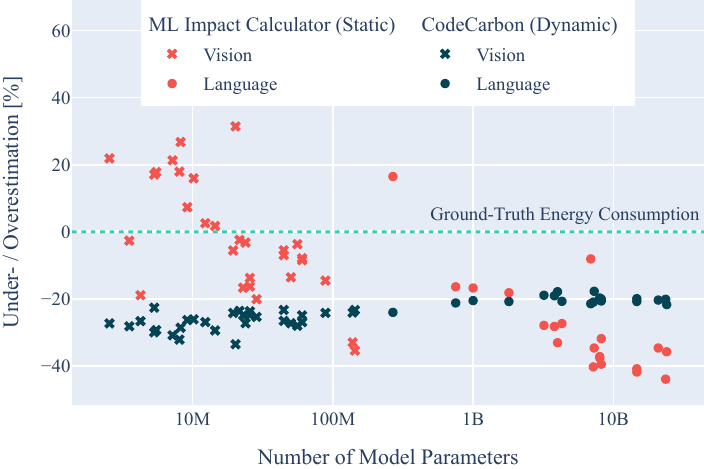}
    \caption{Compared to ground-truth energy measurements, the \emph{ML Emissions Calculator}~\cite{luccioni_quantifying_2019} and \emph{CodeCarbon}~\cite{benoit_courty_2024_11171501} often under- or overestimate the resource demand of AI.}
    \label{fig:opener}
\end{figure}

In order to use and advance AI in a sustainable manner, it is therefore imperative to balance predictive capabilities with resource consumption~\cite{fischer_metaqure,fischer_xpcr}, or put differently, acknowledge and navigate the ``Pareto front'' of ``multi-dimensional model performance''~\cite{fischer_diss,varoquaux_hype_2024}.
Various works explored such trade-offs by investigating the energy consumption and carbon emissions of using AI for generative purposes~\cite{luccioni_power_2024} and classic learning tasks, such as computer vision~\cite{garcia-martin_estimation_2019,schwartz_green_2020,fischer_unified_2022} or natural language processing~\cite{strubell_energy_2020,luccioni_estimating_2023}.
Because respective studies require ``systematic and accurate measurements'' of compute demand~\cite{henderson_towards_2020}, the community developed tools for quantifying the energy consumption of ML.
Among the first projects is the \emph{ML Emissions Calculator}~\cite{luccioni_quantifying_2019}, which statically estimates energy consumption and resulting CO$_2$-equivalents based on user-provided information about the performed experiments.
Shortly later, the \emph{experiment-impact-tracker} was published as a straightforward drop-in library for dynamically tracking compute utilization and estimating carbon emissions of running Python code~\cite{henderson_towards_2020}.
Since the tool was not developed much further, it was quickly outshined by \emph{CodeCarbon}~\cite{benoit_courty_2024_11171501}, a library that does not only allow for integrating resource estimations into Python code, but moreover offers an interactive dashboard view and useful emission comparisons (e.g., household consumption or car driving).
All tools became widely adopted by resource-aware practitioners, with the \emph{ML Emissions Calculator} being cited over 1000 times on \emph{Google Scholar}~\cite{luccioni_quantifying_2019} and \emph{CodeCarbon} surpassing 1500 stars on \emph{GitHub} and nearly 2 million downloads on \emph{PyPI}~\cite{benoit_courty_2024_11171501}.

\begin{figure*}
    \centering
    \includegraphics[width=.84\linewidth]{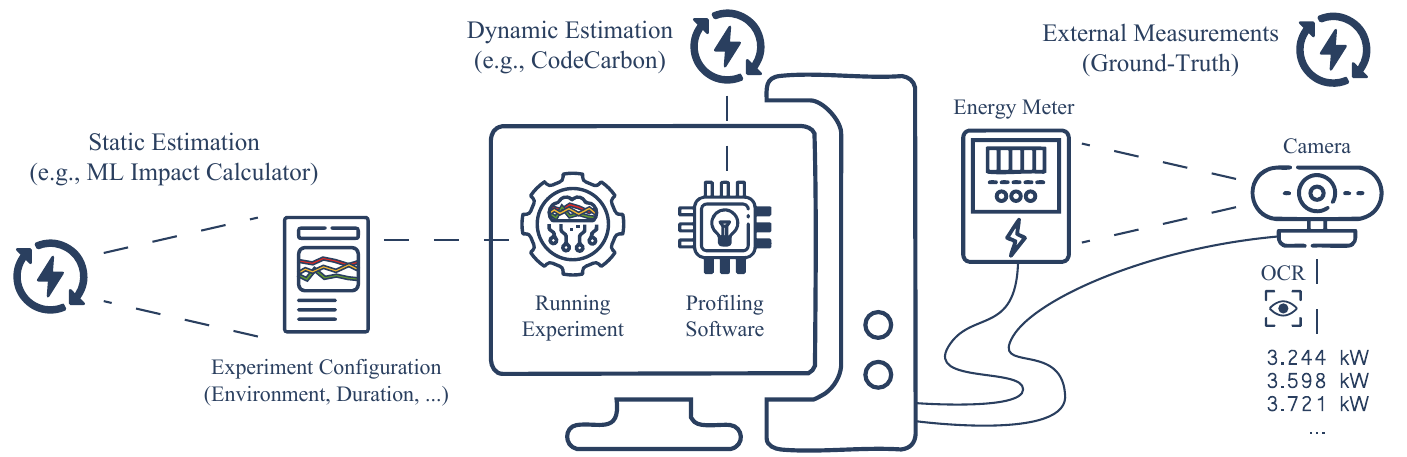}
    \caption{Schematic visualization for validating the static and dynamic estimation of energy demand via external measurements.}
    \label{fig:framework}
\end{figure*}

While the aforementioned initiatives are without a doubt valuable contributions for sustainable AI development, a crucial question remains: \textbf{How accurately do they estimate the energy consumption of AI?}
Static approaches like the \emph{ML Emissions Calculator} assume a constant power draw of the hardware, which however might change across different experiments.
Dynamic estimations via \emph{CodeCarbon} consider the actual compute utilization, however neglect hardware components that cannot be profiled from software, such as the power supply unit, cooling overhead, or peripheral devices.
Figure \ref{fig:opener} summarizes these issues as a teaser to later investigations, with each point representing a single AI experiment.
In comparison with ground-truth measurements (turquoise), both the static (salmon) and dynamic (teal) tools under- or overestimate the energy consumption of most experiments by up to 40\%.
This motivates the study at hand, which sheds light on the unknown figures of AI resource consumption via four key contributions:

\begin{itemize}
    \item Methodology for validating static and dynamic resource estimation approaches (see Figure \ref{fig:framework})
    \item Code base for reproducing and extending the experiments to other domains and estimation approaches
    \item Experimental results that explore AI resource consumption and estimation errors
    \item Discussion on related works, takeaways, limitations, and implications for future research
\end{itemize}

By investigating the ground-truth energy demand of 50 models from the vision and language domain, this work fills an important gap in sustainable AI literature.
It provides empiric evidence that popular tools like \emph{CodeCarbon} and the \emph{ML Emissions Calculator} should be enhanced to account for the missing factors in their internal estimation procedures.
These insights empower researchers and practitioners to make informed decisions about the tools and methods they use for sustainability assessments, carbon accounting, and energy-aware AI system design and deployment.

\section{Methodology}

The core concepts for validating the estimation of AI energy consumption are schematically summarized in Figure \ref{fig:framework}, guiding this section.
Practitioners who want to assess the environmental impacts of using or developing AI generally have three ways for quantifying respective numbers, to be introduced next.
The terminology was aligned with the framework for sustainable and trustworthy reporting~\cite{fischer_diss,fischer_towards_2024}---AI \emph{experiments} (i.e., evaluations) are \emph{executed} in an \emph{environment}, which represents the hardware and software.
The experiment is characterized by a \emph{configuration}, which describes the learning task, dataset, and AI model (or method) at hand.
While this work focuses on AI deployment (i.e., inference with pre-trained models), the following can be easily applied to other tasks.

\subsection{Static Estimation}
At the most abstract level, the configuration and environment of any executed experiment can be utilized to statically estimate the corresponding energy demand and resulting environmental impact.
This approach is also implemented by the \emph{ML Impact Calculator}~\cite{luccioni_quantifying_2019}, which estimates the impact via the total amount of CO$_2$-equivalents:

\begin{equation}\label{ml_impact}
    \texttt{CO}_\texttt{2}\texttt{-Equiv} =
    \texttt{Power} \cdot \texttt{Time} \cdot \texttt{CO}_\texttt{2}\texttt{ Efficiency}
\end{equation}

All of the factors are constants derived from the given configuration and environment, with the power consumption in Watt commonly reflecting the main processor's thermal design power (TDP) or processor base power (PBP), resulting in $\texttt{Energy} = \texttt{Power} \cdot \texttt{Time}$. 
This pragmatic simplification understands the main processor as the biggest factor for the overall energy demand of running AI, which is in line with respective studies~\cite{9005632}, however hardly reflects the complexity of modern execution environments.
The $\texttt{CO}_\texttt{2}\texttt{ Efficiency}$ describes the amount of CO$_2$ emitted per unit energy, which is primarily subject to the local energy mix~\cite{luccioni_quantifying_2019}.
Static estimations only require to measure the running time and are thus easiest to perform, however fail at accounting for dynamic compute utilization and hardware complexity.
To give some examples for non-captured phenomena, running models of varying size will affect the power draw, performing different tasks (e.g., training or inference) might result in differences of resource demand, and compute utilization of a single experiment could even change over time.
In short, while serving practitioners with a straightforward solution to report on the environmental impacts of running AI, static estimations remain rather vague approximations.

\subsection{Dynamic Estimation}

The second estimation approach aims at accounting for dynamic resource demand, thus making the $\texttt{Power} \times \texttt{Running Time}$ in Equation \ref{ml_impact} less static.
In practice, this requires to run additional \emph{profiling software} on the execution environment at hand, such as the aforementioned \emph{experiment-impact-tracker}~\cite{henderson_towards_2020} or \emph{CodeCarbon}~\cite{benoit_courty_2024_11171501}.
Internally, these libraries profile the power consumption of the central or graphics processing unit (CPU / GPU)  via manufacturer tools such as \emph{Intel}'s running average power limit~\cite{intel_sdm_vol3_2024} and \emph{NVIDIA}'s management library~\cite{nvml}.
They allow to iteratively capture the current processor power draw at any point in time ($t$).
By performing respective measurements in short temporal intervals ($\Delta_{\texttt{Time}} = t-(t-1)$), it is possible to estimate the resulting energy draw over time, i.e., $\texttt{Energy} = \sum_{t=1}^T \texttt{Power}_t \cdot \Delta_{\texttt{Time}}$.

Unfortunately, few hardware systems have built-in capabilities for tracking the power consumption of other components than processors, neglecting for example the resource demand of peripheral devices, power supply, and cooling.
The overhead impact of the latter recently received more attention, with studies revealing that cooling can attribute to over 10\% of the overall AI power consumption~\cite{10.1145/3679240.3734609,latif2025coolingmattersbenchmarkinglarge}.
Moreover, special adaptions are required for profiling specialized hardware devices, such as edge accelerators by \emph{Intel} and \emph{Google}~\cite{staay_stress_testing_2024}, or embedded computing boards (e.g, \emph{NVIDIA Jetson})~\cite{fischer_metaqure}.
It should also be noted that internal hardware profiling usually requires administrative access rights to the system and can cause a marginal energy draw overhead.
To summarize, dynamic estimation solutions are easy to incorporate and likely better capture the actual resource consumption of the system, however face limitations with regard to supported hardware and underestimation.

\subsection{Validation via External Measurements}

\begin{figure*}
    \centering
    \includegraphics[width=.84\linewidth]{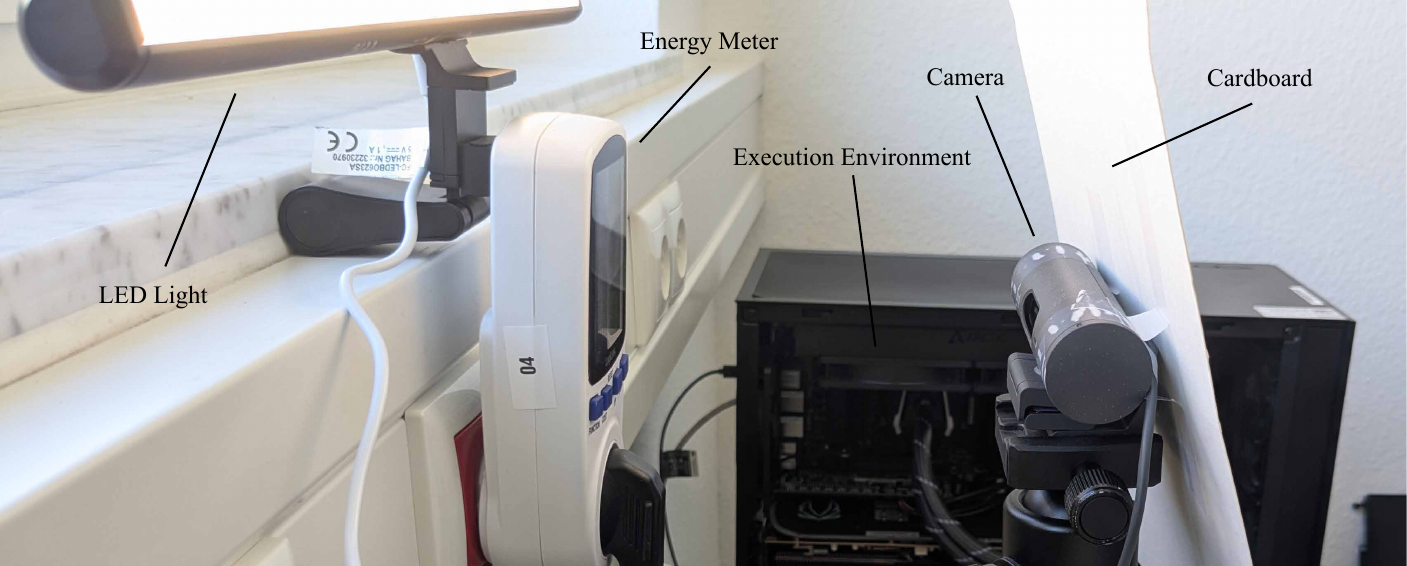}
    \caption{Hardware setup for validating energy estimation with ground-truth data from an energy meter, tracked with a camera.}
    \label{fig:setup2}
\end{figure*}

Overcoming the aforementioned limitations requires to somehow measure the ground-truth $\texttt{Energy}$ consumption.
In order to capture the impact of all environment components, such measurements ideally need to be taken from outside of the system (i.e., externally).
The ground-truth measurements discussed in this work were obtained from a simple yet effective approach, using a standard energy meter, a basic camera, and some straightforward computer vision code.
This setup is easy to realize and can thus be translated to many AI experiment scenarios.

Energy meters exist in various forms, from traditional analogue devices to smart gadgets~\cite{ElectricityMeteringSystems}.
Basic variants are not only highly affordable (less than 10€), but can also be easily installed between the power socket and plug of the hardware on which the AI experiment is executed.
While such simple meters do not usually allow for digitally extracting the measurements, the display can be easily tracked with a standard camera.
This setup allows to capture images of the displayed energy consumption at the start and end of any experiment, which can later be processed to numeric data via basic computer vision and optical character recognition (OCR).

Figure \ref{fig:framework} visualizes how all three approaches can be run in parallel, using only one execution environment equipped with some basic additional devices.
By storing and aggregating the respective logs, it is possible to compare and validate static estimation, dynamic estimation, and external measurements of AI energy consumption.
It is important to note that external measurements also have certain drawbacks---most importantly, they require additional hardware and implementation effort.
In fact, all three approaches have their pros and cons, with this work exploring their relation and complementary nature (see also the later Discussion).

\subsection{Practical Setup}

To empirically compare the different approaches for quantifying AI energy consumption, the introduced concepts were also practically implemented.
Figure \ref{fig:setup2} showcases the experimental setup, using a basic energy meter by \emph{LogiLink}, a camera by \emph{Logitech}, and a deep learning workstation equipped with an \emph{Intel i9-13900K} CPU and \emph{NVIDIA RTX 4090} GPU.
To allow for overnight experiments and alleviate display reflections, the setup also uses an LED light and a white piece of cardboard.
The experiment software was implemented in Python and can be found at \url{https://github.com/raphischer/ai-energy-validation}, which also includes the logs and results.
The dynamic estimate was performed with \emph{CodeCarbon 3.0.1}~\cite{benoit_courty_2024_11171501}, while the static estimate assumed a TDP of 300 Watt for the GPU (taken from the \emph{ML Impact Calculator}~\cite{luccioni_quantifying_2019}) and 125 Watt for the CPU (constant in \emph{CodeCarbon}'s $\texttt{cpu\_power.csv}$ file).
As additional dependencies, experiment execution was streamlined with \emph{mlflow}~\cite{Zaharia_Accelerating_the_Machine_2018}, camera access and computer vision was enabled by \emph{opencv-python}~\cite{opencv_library}, OCR was performed with \emph{scikit-learn}~\cite{scikit-learn} (using a custom random forest classifier), the logs were analyzed with \emph{pandas}~\cite{team_pandas-devpandas_2020}, and all plots were created with \emph{plotly}~\cite{plotly_technologies_inc_plotly_2024}.

Because AI deployment quickly exceeds the energy demand of training, the experiments focused on the inference performance of pre-trained AI models.
To validate AI energy consumption across different configurations, the performance of 30 image classifiers~\cite{simonyan_very_2015,szegedy_rethinking_2016,he_deep_2016,howard_mobilenets_2017,he_identity_2016,huang_densely_2017,chollet_xception_2017,sandler_mobilenetv2_2018,zoph_learning_2018,howard_searching_2019,tan_efficientnet_2019,InceptionResNetV2,EfficientNetV2,ConvNeXt_22}, available from \emph{Keras 3}~\cite{chollet2015keras}, as well as 20 large language models~\cite{jiang2023mistral7b,zhang2024tinyllamaopensourcesmalllanguage,grattafiori2024llama3herdmodels,abdin2024phi4technicalreport,abdin2024phi3technicalreporthighly,olmo20252olmo2furious,deepseekai2025deepseekr1incentivizingreasoningcapability,gemmateam2025gemma3technicalreport,mistralai2025magistral,yang2025qwen3technicalreport,openai2025gptoss120bgptoss20bmodel} offered by \emph{Ollama 0.11.8}~\cite{ollama} was measured.
Later referred to as \emph{Vision} and \emph{Language}, these open weights models represent examples from two popular deep learning application domains.
All models were published between 2015 and 2025 and have a complexity between 2 million and 23 billion parameters, making them feasible for single-GPU deployment (i.e., on-premise AI instead of AI-as-a-service~\cite{lins_artificial_2021}).
To assess their performance, the vision models were configured to perform image classification on \emph{ImageNet} samples~\cite{imagenet} for 2 minutes, testing inference on the CPU or GPU with a batch size of 4 or 16 (i.e., processing 4 or 16 images at a time).
The language models were confronted with random prompts from the \emph{Puffin} dataset~\cite{ldjnr_puffin_2023} for 15 minutes, using a temperature of 0.1 or 0.7 (lower value makes the models behave more deterministic with regard to token probabilities).
If not otherwise specified, the following displays results for the default batch size of 16 and temperature of 0.7.
All results were averaged over three consecutive runs that reduce random outlying behavior, with standard deviation information indicated by error bars.

Running all experiment runs took less than three days and resulted in carbon emissions based on the consumed energy, connecting back to quantifying environmental impacts~\cite{luccioni_quantifying_2019} via Equation \ref{ml_impact}.
With a ground-truth energy consumption of $20.04 \texttt{ kWh}$, and assuming a $\texttt{CO}_\texttt{2}\texttt{ Efficiency}$ of 0.38 for Germany~\cite{zeppelin2024conversion}, the total emissions of the experiments are estimated to $20.04 \texttt{ kWh} \cdot 0.38 \texttt{ kgCO}_\texttt{2}\texttt{/kWh} \approx 7.62\texttt{ kg CO}_\texttt{2}\texttt{-Equiv}$ (not taking into account any embodied factors~\cite{falk2025carboncradletograveenvironmentalimpacts} or overhead efforts from development, testing, and writing).
Keep also in mind this work only focuses on model performance in terms of resource consumption and does not explore any efficiency trade-offs with prediction quality~\cite{fischer_diss}.

\section{Results}

\begin{figure*}
    \centering
    \includegraphics[width=.84\linewidth]{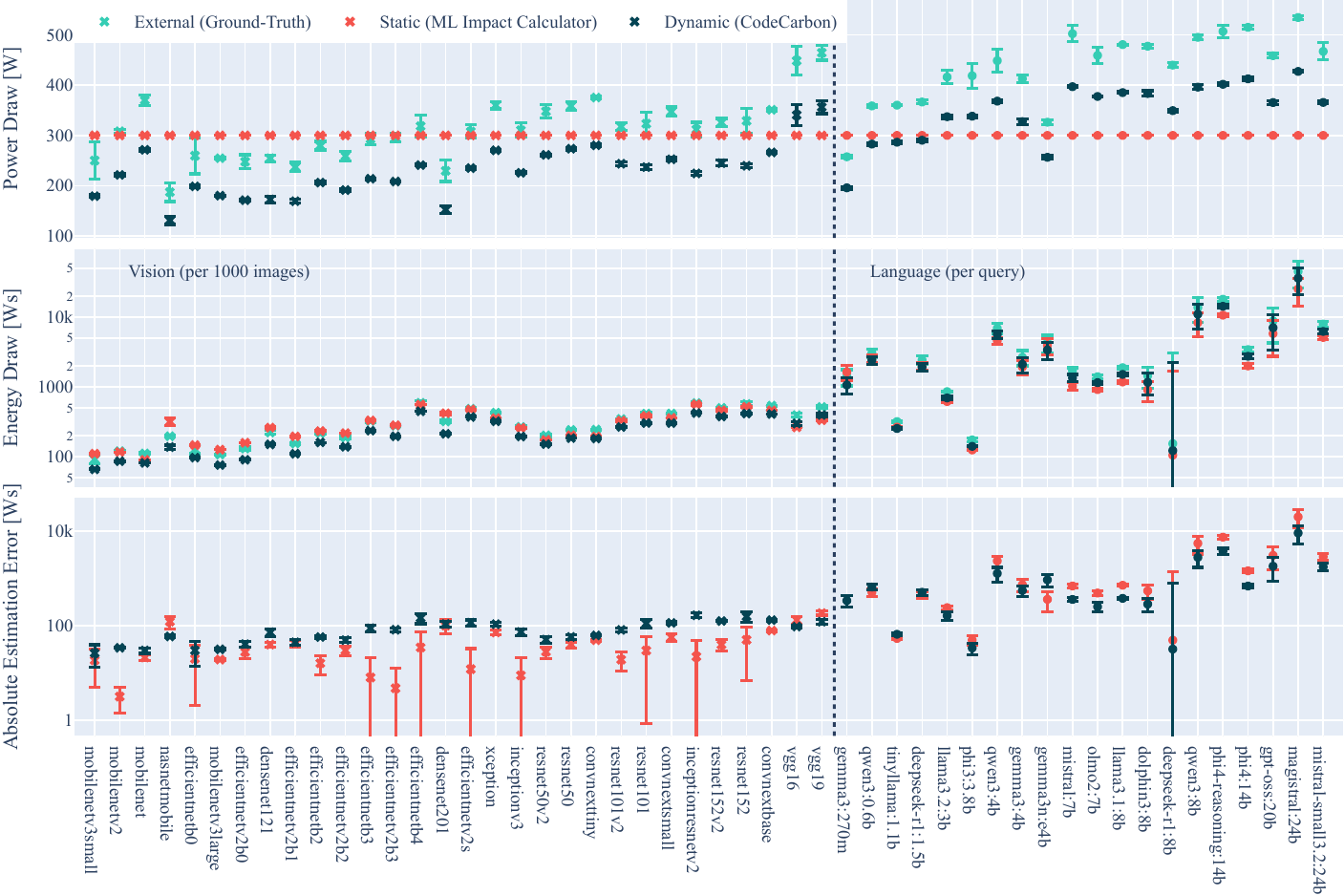}
    \caption{Power draw of the execution environment (first row), energy draw for performing inference with a fixed number of samples (second row), and energy estimation errors (third row), assessed for all three approaches and across all tested models. There is a clear upwards trend of power, energy, and error, generally growing with model complexity. The energy estimation approaches follow the general patterns of the ground-truth data but make significant estimation errors.}
    \label{fig:groundtruth_power}
\end{figure*}

Recall that a high-level summary of the experimental results was already given in Figure \ref{fig:opener}, however the following explores them in more depth.
To investigate the alignment of energy estimations and ground-truth data, the first row of Figure \ref{fig:groundtruth_power} displays the average power draw (Watt, $y$-axis) of all tested models ($x$-axis) during the experiment.
The models were sorted based on their parameter count, from \emph{mobilenetV3small} (2.5m parameters)~\cite{howard_searching_2019} to \emph{mistral-small3.2:24b}~\cite{mistralai2025magistral}, resulting in a visible upwards trend of power consumption due to higher compute utilization.
Nevertheless, there is clear evidence that the power draw does not only depend on the parameter count---some of the models are more power-efficient than expected from their complexity, such as \emph{nasnetmobile}~\cite{zoph_learning_2018}, \emph{densenet201}~\cite{huang_densely_2017}, and the \emph{gemma3} variants~\cite{gemmateam2025gemma3technicalreport},
Looking at the different curves, we also see that the constant value assumed by the static energy estimation (salmon points at 300 Watt) differs from the ground-truth measurements (turquoise), especially for the smallest and largest models.
The dynamic profiling (teal) follows the general upwards trend, however compared to the externally measured data, consistently underestimates the power draw.
The absolute difference between the measurements and estimates often exceeds 100 Watt, which is a significant error considering the TDP of 300 Watt~\cite{luccioni_quantifying_2019},
Impressively, the maximum observed power draw (534 Watt for \emph{magistral:24b}~\cite{mistralai2025magistral}) is nearly twice as high as the static estimate.

In the second row of Figure \ref{fig:groundtruth_power}, we see the amount of energy (Watt-seconds, logarithmic $y$-axis) required to classify 1000 images (vision) or answering a single query (language).
Once again showcasing an upwards trend, this demonstrates the considerably higher resource demand of generative language models compared to image classifiers~\cite{luccioni_power_2024}.
While all investigated models belong to the family of deep neural networks and consist of many million parameters, the classifiers are clearly faster and more energy-friendly, processing many thousand images in the time that language models need for answering a single prompt.
As before, the energy demand does not strictly increase with parameter count---some models like \emph{tinyllama:1.1b}~\cite{zhang2024tinyllamaopensourcesmalllanguage} and \emph{phi3:3.8b}~\cite{abdin2024phi3technicalreporthighly} answer prompts with much less energy than expected from the complexity.
We moreover see that both estimation approaches are generally able to follow the rough patterns of energy consumption, however also make errors.
The static estimates are too high for small models and too low for bigger models, while the dynamic approach always underestimates the energy demand. 
One should note that these differences are expected to increase when running the models for longer periods of time, as the constant mismatch of power consumption (first row) will increase the gap over time.

The absolute energy estimation errors are displayed in the last row of Figure \ref{fig:groundtruth_power}, connecting back to the relative results in Figure \ref{fig:opener}.
The absolute values were used in order to also display the errors on a logarithmic scale. 
For smaller classifiers, the static estimation results in lower errors (note that some of them actually represent an overestimation), while for the bigger models, the dynamic estimates are more accurate.
This can also be observed in the other rows, where the teal (dynamic) data ''overtakes`` the salmon (static) estimations with growing model complexity.
As we once again find an upwards trend in the bottom row, we also see evidence that the estimation error tends to grow with model size and resource demand.
Estimation errors below 10 Watt-seconds are extremely rare and could only be observed for four of the vision models.
Across all three rows of the plot, the standard deviations across the experimental runs are generally rather low, demonstrating stability of the results with respect to randomization effects---except for \emph{deepseek-r1:8b}~\cite{deepseekai2025deepseekr1incentivizingreasoningcapability}, for which strong variations across different queries are observed.

\begin{figure}[t]
    \centering
    \includegraphics[width=.84\linewidth]{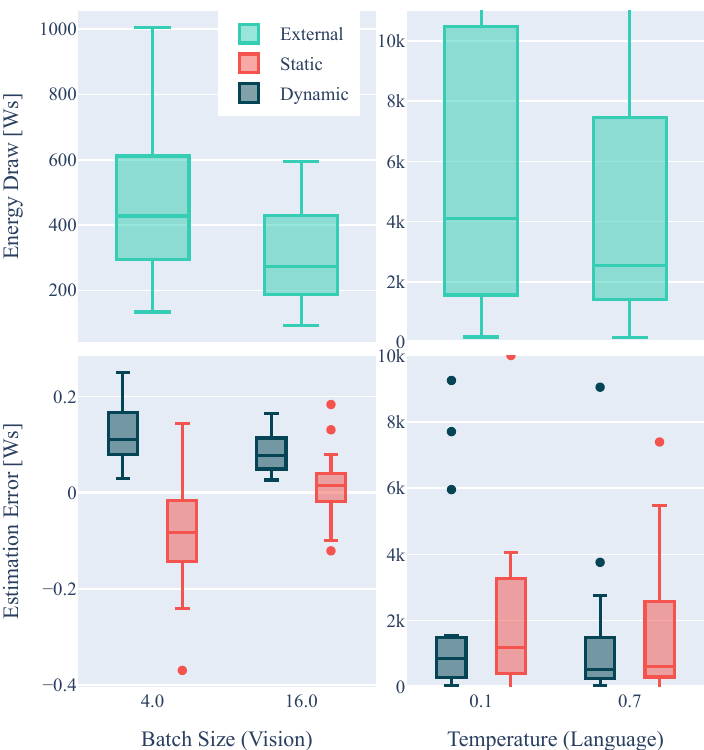}
    \caption{Impact of hyperparameter choice ($x$-axes) for AI energy demand. The higher batch size and temperature results in more efficient compute utilization (i.e., lower energy draw) for vision and language models, which also lead to lower estimation errors.}
    \label{fig:hyperparameter_impact}
\end{figure}

As another point of this experimental investigation, we next explore how the batch size and temperature configuration impacts the resource consumption of vision and language models, respectively.
Figure \ref{fig:hyperparameter_impact} showcases the ground-truth energy demand (top row) as well as the energy estimation errors (bottom row) when configuring the models with these hyperparameters and once more classifying 1000 images or answering a single query---the resulting performance distribution is summarized via box plots.
One should note that the image batch size choice (left) only affects the running time and resource demand of the vision models, however does not impact their prediction quality.
The results demonstrate that larger batches result in lower energy draw and thus also reduce the static and dynamic estimation errors across all models, however the batch size is naturally capped by the GPU memory capacity (i.e., larger batch sizes would not have been feasible for some of the more complex models).
The temperature (right) actually affects the language model output, as it weights the predicted token probabilities to control how explorative or deterministic the provided query is answered.
We see that the lower temperature (i.e., higher focus on token probabilities) leads to higher energy demand and larger estimation errors across the models.
However, this configuration also seems to introduce more variation, resulting in considerably larger boxes.
These results evidence that hyperparameters can have a strong impact on the resource demand of AI models, necessitating to explicitly report on the specific configuration of evaluated models.

\begin{figure}[t]
    \centering
    \includegraphics[width=.84\linewidth]{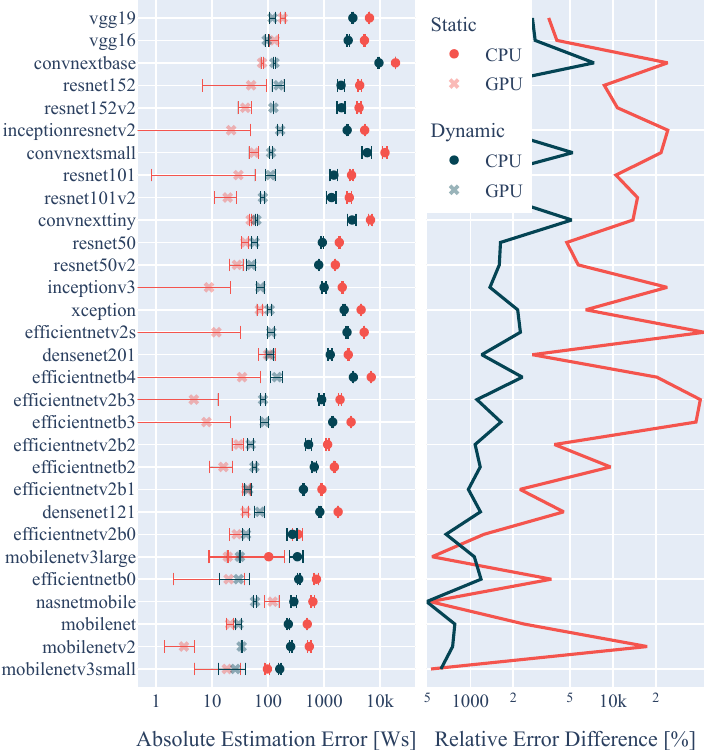}
    \caption{Absolute energy estimation errors of vision models when deployed on the CPU or GPU (left). Both the static and dynamic approaches make higher errors for CPU inference, and the differences between the CPU and GPU errors (right) grow with model complexity.}
    \label{fig:cpu_vs_gpu}
\end{figure}

While deep learning models have become especially popular thanks to the efficiency of GPU deployment, there are also cases where practitioners are restricted to running AI on a CPU.
This raises the question of whether energy estimation inaccuracies are affected by performing inference on the CPU or GPU.
For this investigation, Figure \ref{fig:cpu_vs_gpu} lists the respective absolute estimation errors (left logarithmic $x$-axis) for the vision models, when using or disabling the GPU---accordingly, the faded scatter points represent the same data displayed in the last row of Figure \ref{fig:groundtruth_power}.
This comparison clearly shows that the errors of both estimation approaches are considerably higher when only utilizing the CPU.
As before, the errors grow with model size (i.e., upwards along the $y$-axis), and the differences between the CPU and GPU errors seem to also increase.
In the most extreme case (\emph{EfficientNetV2s}~\cite{EfficientNetV2}), the dynamic estimation error for CPU inference (5225 Ws) is over 463 times as high as the respective GPU error (12 Ws).
On the right hand side, the difference between the CPU and GPU estimation errors is shown more explicitly, in relation to the GPU errors.
It demonstrates that the relative CPU error also increases with model complexity and is more than ten times (1000\%) higher for nearly all configurations.
As before, variations can be observed across the models---the \emph{ConvNeXt}~\cite{ConvNeXt_22} and \emph{EfficientNet}~\cite{tan_efficientnet_2019} variants have the highest error differences, while some models with comparable parameter count show less deviations.
Moreover, the static estimation error differences (using a TDP of 125 Watt for the CPU) are considerably higher than the ones observed for dynamic estimations (except for \emph{MobileNetV3Large}~\cite{howard_searching_2019}).
As such, the results indicate that the choice of execution environment and processor strongly impacts the accuracy of energy estimation tools, expecting higher errors for CPU-only inference.

\section{Discussion}

Let us interpret the experimental findings in the context of related literature.
Various works have investigated AI energy consumption with the help of \emph{CodeCarbon}~\cite{luccioni_estimating_2023, luccioni_power_2024,staay_stress_testing_2024,fischer_diss}, however our findings indicate that their reported numbers are underestimations.
The results also underline the importance of acknowledging the overhead impact of cooling~\cite{10.1145/3679240.3734609,latif2025coolingmattersbenchmarkinglarge}, which is one of the factors neglected by static and dynamic energy estimation approaches.
It should be noted that holistic reporting on the environmental impacts of AI requires to investigate more factors of the AI life cycle~\cite{wu_sustainable_2022}, including intricate phenomena like embodied impacts~\cite{falk2025carboncradletograveenvironmentalimpacts} and rebound effects~\cite{10.1145/3715275.3732007}.
Nevertheless, I believe that reporting on the ground-truth energy consumption of running AI models remains a crucial ingredient for advancing the field in environmentally sustainable ways~\cite{fischer_towards_2024}.
To that end, the investigations of this study can be useful for refining established tools, for example by incorporating constant factors that account for estimation errors or featuring explicit disclaimers about inaccuracies.
From the experimental analysis, six important \textbf{takeaway points} can be formulated:

\begin{enumerate}
    \item Energy consumption generally scales with model complexity and size, however certain AI architectures use their parameters more efficiently than others
    \item Energy estimation approaches likely resulted in inaccurate reportings of environmental impacts, featuring errors that grow with model size and time
    \item Static estimation hits a sweet spot with low errors for mid-sized models, but under- or overestimates the demand of large or small architectures by -40 to 40 \%
    \item Dynamic estimation better follows the ground-truth patterns, but consistently underestimates the consumption by 20--30\%
    \item Hyperparameters like batch size or temperature impact the efficiency and estimation errors, and accordingly should be carefully tuned and reported 
    \item Estimating the energy demand of CPU-only deployment results in errors that are considerably higher than for GPU inference
\end{enumerate}

While providing important insights, this study also faces limitations, such as only evaluating a single environment.
Other works have already evidenced that the choice of hardware and software impacts AI resource efficiency~\cite{fischer_unified_2022,fischer_metaqure}, and this investigation should be seen as a first step toward more reliable energy estimations.
The presented concepts and accompanying code repository allow to easily extend the investigation to other learning domains and evaluation setups.
Future work could collect more data and deepen the analysis with regard to relations between configurations, environments, and resulting estimation errors.
A second limitation lies in the external measurement approach, which might not be applicable to all AI deployment scenarios---this actually also holds true for static and dynamic estimations.
For example, the energy consumption of compute clusters and data centers can be hardly monitored with basic energy meters, requiring different solutions for measuring ground-truth data, such as power distribution units with outlet level metering.
Nevertheless, the proposed validation framework is applicable and affordable for on-premise deployment scenarios, which remains popular for the sake of privacy.
What also remains for future work is a deeper investigation of how resource consumption trades with predictive quality~\cite{fischer_metaqure} or other performance aspects~\cite{fischer_xpcr}, as past studies have focused on vision models~\cite{fischer_unified_2022}.
In closing this discussion, I would like to emphasise once more that I fully support the aforementioned energy estimation initiatives---their published tools are invaluable resources for environmentally-aware practitioners, and I hope that my work will contribute to their continued improvement.

\section{Conclusion}

To summarize, this study validated energy estimation tools such as \emph{CodeCarbon} and the \emph{ML Impact Calculator}, which succeed at capturing the rough resource consumption of evaluated AI models.
However, comparing their quantified numbers with externally measured ground-truth data also revealed significant estimation errors.
Growing model complexity was observed to increase AI energy demand and estimation errors, which can even be further impacted by choice of hyperparameters and processor.
With this work, resource-aware practitioners are equipped with means for investigating the actual energy consumption of their local AI experiments.
Moreover, developers of energy estimation tools can improve their solutions along the guidelines of this work, accounting for estimation errors of their approaches.
As such, the presented contributions establish transparency and not only promote but facilitate the sustainable development and use of AI.

\printbibliography

\end{document}